\documentclass[conference]{IEEEtran}
\IEEEoverridecommandlockouts
\usepackage{cite}
\usepackage{amsmath,amssymb,amsfonts}
\usepackage{algorithmic}
\usepackage{graphicx}
\usepackage{textcomp}
\usepackage{xcolor}
\usepackage{comment}
\usepackage{booktabs} 
\usepackage{multirow} 
\usepackage[margin=1in]{geometry}
\usepackage[most]{tcolorbox}
\usepackage{enumitem}
\usepackage{microtype}
\usepackage{lmodern}

\definecolor{PromptBlue}{HTML}{EDF5FC}
\definecolor{BorderBlue}{HTML}{315B83}
\definecolor{HeadingBlue}{HTML}{244A70}
\usepackage{tikz}
\usetikzlibrary{arrows.meta, positioning, fit}
\usepackage{xcolor}

\def\BibTeX{{\rm B\kern-.05em{\sc i\kern-.025em b}\kern-.08em
    T\kern-.1667em\lower.7ex\hbox{E}\kern-.125emX}}
\begin{document}

\title{LLM-Assisted Label Noise Correction for EEG-Based Sleep Stage Classification\\

}

\author{
\IEEEauthorblockN{Elmira Salari, Shruti Kshirsagar}
\IEEEauthorblockA{
\textit{School of Computing} \\
\textit{Wichita State University} \\
Wichita, Kansas, USA \\
exsalari1@shockers.wichita.edu
}
}
\maketitle

\begin{abstract}
Automatic sleep stage classification from electroencephalography (EEG) signals is a critical task in clinical sleep medicine. While deep learning models have shown promising results in this domain, their performance is fundamentally constrained by the quality of training labels, which human experts manually annotate and that are subject to inter-rater disagreement and systematic labeling errors. In this paper, we propose a novel LLM-assisted label-noise filtering pipeline that operates exclusively during training to identify and remove unreliable sleep-stage annotations before retraining the classifier. Specifically, TinySleepNet is first trained on the original dataset to extract per-epoch softmax confidence scores, then low-confidence epochs are submitted to LLaMA 3.1 along with their EEG spectral features and surrounding sequence context. The LLM evaluates each epoch against AASM sleep-scoring guidelines and assigns a verdict of \textit{RELIABLE} or \textit{SUSPICIOUS}. Suspicious epochs are removed from the training set, and TinySleepNet is retrained on the cleaned data. Evaluated on the Sleep-EDF Expanded dataset using 20-fold cross-validation, our method improves overall accuracy from 81.8\% to 84.6\% and Macro F1-score from 75.6\% to 77.4\%, with particularly notable gains in REM detection. Crucially, because the LLM operates only during training, the lightweight TinySleepNet architecture (1.3M parameters) is preserved at inference, making the proposed pipeline suitable for real-time, wearable sleep monitoring applications.
\end{abstract}


\section{Introduction}
Sleep staging is a fundamental component of sleep analysis, enabling the diagnosis of sleep disorders and assessment of sleep quality \cite{b1, b2}. Traditionally, clinical experts perform sleep staging using polysomnography (PSG), including the electroencephalogram (EEG) and other physiological signals, along with sleep staging rules. Depending on which sleep-staging standard is used, the number of stages may vary. The American Academy of Sleep Medicine (AASM) sleep staging standard divides each 30-second sleep epoch into 5 stages: W, N1, N2, N3, and REM \cite{b3}. However, manual sleep staging is a time-consuming, labor-intensive task and is subject to significant inter-rater variability, especially for transitional stages such as N1 and the boundary between N2 and N3 \cite{b4}. This variability introduces label noise into training datasets. According to \cite{b5}, inter-rater agreement for sleep staging can be as low as 72-82\% even among experienced clinicians.

Although Machine Learning (ML) and Deep Learning (DL) approaches have made significant progress in automated sleep staging \cite{b6,b7,b8,b9}, these models are all trained on potentially noisy human-annotated labels, and the impact of label noise on their performance has received limited attention.

The advent of Large Language Models (LLMs) has enabled the incorporation of clinical domain knowledge into data processing pipelines \cite{b10} because these models are trained on large-scale medical corpora. Recent studies further explore their integration with physiological data, including EEG, for applications such as sleep stage analysis and classification \cite{b11}. This suggests these models could serve as complementary tools for assessing the consistency between EEG-derived features and human-assigned sleep stage labels.

In this work, we propose a novel approach that uses an LLM as a training-time label noise filter. Our key insight is to use the LLM only during the data-cleaning phase, rather than replacing the lightweight sleep-staging classifier with a heavy LLM, thereby preserving inference-time computational efficiency.

Our main contributions are as follows:
\begin{itemize}
    \item We propose the first LLM-based training-time label-noise filter for EEG sleep staging, which operates without modifying the inference pipeline.
    \item The LLM produces human-readable reasoning for each filtering decision, making the data cleaning process interpretable and clinically auditable.
\end{itemize}

\section{Related Work}
\subsection{Automated Sleep Stage Classification}
Sleep stage classification has evolved significantly over the decades, transitioning from manual scoring to fully automated approaches. Early ML methods such as SVM \cite{b6} and Random Forests \cite{b12} showed promising results but relied heavily on handcrafted features that required domain expertise. DL approaches subsequently addressed this limitation by learning representations directly from raw EEG signals. For example, Supratak et al. introduced DeepSleepNet \cite{b7}, combining CNNs for feature extraction with bidirectional LSTMs for sequence modeling, achieving state-of-the-art results on the Sleep-EDF dataset \cite{b14}. Supratak and Guo subsequently 
proposed TinySleepNet \cite{b15}, which is a compact variant with 1.3M parameters that maintains competitive accuracy while being suitable for resource-constrained deployment. Phan et al. introduced SleepTransformer \cite{b9}, applying transformer architectures to model long-range sequences of epochs.

 Despite the significant performance of some of the mentioned approaches, these models are purely data-driven and require large amounts of accurately labeled training data, making them sensitive to label noise introduced by inter-rater disagreement during manual annotation, the limitation that motivates our work.

\subsection{Label Noise in Medical Data}
Label noise, which relates to the presence of incorrect or inconsistent labels in training data, is a well-recognized challenge in supervised learning \cite{b17}.
In medical domains, this problem is more pronounced because of the subjective nature of expert annotation and the inherent ambiguity of clinical criteria.
In sleep staging specifically, inter-rater agreement among expert scorers has been reported to range from 72\% to 82\% \cite{b16, b17, b18}. This variability introduces systematic label noise into training datasets, which might negatively impact the performance of supervised DL models, particularly for minority and transitional sleep stages.

Several approaches have been proposed to address label noise in general ML settings. For instance, Northcutt et al. \cite{b19} proposed Confident Learning, a framework for characterizing and correcting label errors by estimating the joint distribution between noisy and true labels. Likewise, co-training \cite{b20} and mixture-of-experts \cite{b21} approaches have also been explored to reduce the impact of noisy labels during training. However, these methods are general-purpose and do not leverage clinical knowledge to assess label reliability. 

\subsection{LLMs in Medical Applications}
LLMs have demonstrated strong performance across a variety of medical tasks, including medical question answering, clinical decision support, and biomedical text analysis \cite{b11,b22}. In the context of biosignal analysis, LLMs have been applied to ECG interpretation \cite{b13} and clinical note generation \cite{b24}. However, few studies examine their use as training-time data-cleaning agents for EEG classification.

The most closely related work is Qiu et al. \cite{b25}, which uses a Multimodal LLM as the final sleep-stage classifier by converting EEG epochs into images and leveraging multi-perspective agreement with self-training to handle label noise. Their approach achieves 85.2\% accuracy on Sleep-EDF; their method requires an 11B parameter model at inference time, a LoRA fine-tuning step, and specialized GPU hardware. In contrast, our method uses the LLM only at training time as a noise filter, keeping the lightweight TinySleepNet \cite{b15} as the inference model. This makes our approach light and practical for real-time clinical scenarios.

\section{Methodology}

\subsection{Dataset}\label{AA}
In this study, we used the Sleep-EDF Database Expanded v1.0.0 Cassette category, which includes 153 PSG recordings from 78 healthy subjects aged 25 to 101 years. Sleep stages are labeled according to the Rechtschaffen and Kales (R\&K) manual \cite{b26}, and then mapped to the five AASM stages: W, N1, N2, N3, and REM. Although the PSG recording includes EEG, EOG, chin EMG, respiration, and body temperature, we used a single-channel EEG (Fpz-Cz) sampled at 100Hz, the most commonly used channel in automatic sleep staging, to enable direct comparison with existing methods. 

We used the same preprocessing approach as in the TinySleepNet repository, extracting the EEG (Fpz-Cz) channel at 100 Hz and segmenting it into 30-second non-overlapping epochs, yielding 3000 samples per epoch. To focus on the sleep period and reduce excessive wake epochs, we retained only epochs within the 30 minutes before the first non-wake epoch and after the last non-wake epoch, and removed epochs labeled as movement or unknown. As mentioned previously, stage labels use the standard 5 classes: N1, N2, N3, W, and REM.

We adapted the standard subject-wise 20-fold cross-validation protocol. To prevent data leakage, recordings from the same participant should be assigned to the same data partition. Due to variations in recording duration and sleep efficiency, the number of epochs varies across folds. Table~\ref{tab:dataset} summarizes the epoch distribution across training, validation, and test sets.

\begin{table}[t]
\centering
\caption{Epoch Distribution Across 20 Cross-Validation Folds}
\label{tab:dataset}
\begin{tabular}{lrrr}
\hline
\textbf{Data Split} & \textbf{Mean} &
\textbf{Min} & \textbf{Max} \\
\hline
Training   & 38,070 & 37,025 & 39,306 \\
Validation & 3,938  & 3,827  & 5,227  \\
Test       & 2,211  & 1,028  & 3,368  \\
\hline
\multicolumn{4}{l}{\textbf{Total Unique Epochs: 195,479}} \\
\hline
\end{tabular}
\end{table}

\begin{figure*}[h]
\centering
\includegraphics[width= 0.8\textwidth]{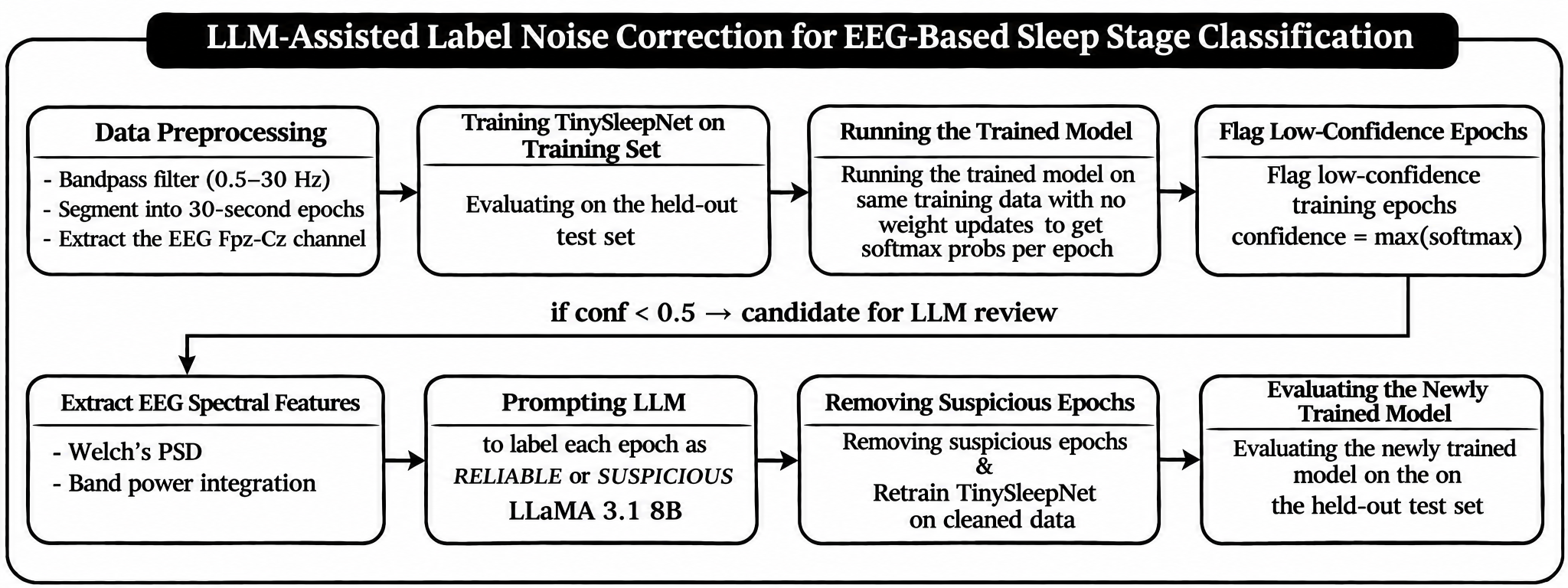}
\caption{Proposed Model Overview}
\label{fig:LLM-Assisted EEG Label Noise Correction Flowchart}
\end{figure*}

\subsection{Baseline Model: TinySleepNet}
We used TinySleepNet as our baseline model due to its efficiency and strong performance in single-channel EEG sleep staging. The model combines a CNN for automatic feature extraction from raw EEG signals with a single-layer LSTM to capture temporal dependencies between consecutive epochs. The model has only 1.3 million parameters, which makes it a lightweight yet competitive design for our approach. 

\subsection{LLM-Assisted Label Noise Approach}
Our proposed pipeline introduces a novel label-noise filtering mechanism that operates exclusively during training. To do so, we leveraged an LLM (LLaMA 3.1 8B) to identify and remove unreliable sleep-stage annotations before retraining the model. The pipeline consists of several main stages, as explained below. The details of our proposed methodology are shown in Fig~\ref{fig:LLM-Assisted EEG Label Noise Correction Flowchart}.

\subsubsection{Confidence-Based Candidate Selection}
After training TinySleepNet on the original dataset, the trained model is run in inference mode on the same training data to obtain per-epoch softmax probability vectors. The confidence score for each epoch is defined as the maximum softmax probability:
\begin{equation}
c_i = \max_{k \in \{W, N1, N2, N3, REM\}} p(k | x_i)
\end{equation} Epochs with confidence below a threshold $\tau$ are selected as candidates for LLM verification. In this study, we used $\tau = 0.5$ as the primary threshold, flagging approximately 22,689 epochs (approximately 3\% of total training data) as candidates.

\begin{figure}[htbp]
\centering

\begin{tcolorbox}[
    enhanced,
    colback=PromptBlue,
    colframe=BorderBlue,
    boxrule=1pt,
    arc=2.5mm,
    width=\linewidth,
    left=12pt,
    right=12pt,
    top=12pt,
    bottom=12pt,
    before upper={\raggedright}
]

\footnotesize
\setlength{\parindent}{0pt}
\setlength{\parskip}{4pt}

You are a sleep scoring expert using AASM guidelines.

A 30-second EEG epoch is labeled
\textbf{[STAGE]} by a human annotator.

Determine if this label is consistent with
the EEG signal features.

\vspace{6pt}
{\color{HeadingBlue}
\textbf{EEG FEATURES:}}

\begin{itemize}[
    leftmargin=12pt,
    nosep,
    topsep=2pt
]
    \item Delta (0.5--4 Hz): [LEVEL] ([VALUE]\%)
    \item Theta (4--8 Hz): [LEVEL] ([VALUE]\%)
    \item Alpha (8--13 Hz): [LEVEL] ([VALUE]\%)
    \item Sigma (11--16 Hz): [LEVEL] ([VALUE]\%)
    \item Beta (16--30 Hz): [LEVEL] ([VALUE]\%)
\end{itemize}

\vspace{6pt}
{\color{HeadingBlue}
\textbf{SURROUNDING EPOCHS:}}

[PREV] $\rightarrow$ [CURRENT] $\rightarrow$ [NEXT]

\vspace{6pt}
{\color{HeadingBlue}
\textbf{CLASSIFIER PREDICTION:}}

\begin{itemize}[
    leftmargin=12pt,
    nosep,
    topsep=2pt
]
    \item Predicted label: [PRED\_LABEL]
    \item Classifier confidence: [SCORE]
    (0=uncertain, 1=certain)
\end{itemize}

\textit{Note: sent for review because
confidence $<$ [THRESHOLD]}

\vspace{6pt}
{\color{HeadingBlue}
\textbf{AASM RULES:}}

[AASM\_RULES]

\vspace{6pt}
{\color{HeadingBlue}
\textbf{DECISION:}}

Is the label \textbf{[STAGE]}
\textbf{RELIABLE} or \textbf{SUSPICIOUS}?

\vspace{5pt}

\textbf{VERDICT:}
[RELIABLE or SUSPICIOUS]

\textbf{LLM\_CONFIDENCE:}
[0.0 to 1.0]

\textbf{REASON:}
[one sentence]

\end{tcolorbox}

\caption{Prompt template for LLM-assisted
sleep-stage label reliability assessment.}
\label{fig:llm_prompt}

\end{figure}

\subsubsection{Spectral Feature Extraction}
For each candidate epoch, we extract five EEG frequency-band powers using the Welch Power Spectral Density (PSD) method. We first filter the signal with a fifth-order Butterworth bandpass filter (0.5--30 Hz), then estimate the PSD using the Welch method with a window size of 256 samples. The power in each band is computed as:

\begin{equation}
P_{band} = \int_{f_{low}}^{f_{high}} S(f) df
\end{equation}
and normalized as a percentage of total power. The five bands are Delta (0.5--4 Hz), Theta (4--8 Hz), Alpha (8--13 Hz), Sigma (11--16 Hz), and Beta (16--30 Hz), selected for their direct correspondence to the AASM sleep scoring criteria.

\subsubsection{LLM-Based Label Verification}
Each candidate epoch is submitted to LLaMA 3.1 
(8B parameters), running locally via Ollama, with a structured prompt containing the spectral features, surrounding epoch context, and classifier prediction. The LLM determines whether the annotator's label is \textit{RELIABLE} or \textit{SUSPICIOUS} based on AASM guidelines, presented features, and neighboring epochs. We also set the temperature to 0.1 to reduce response variability. The Prompt used for this study is depicted in Fig~\ref{fig:llm_prompt}. 
 
\subsubsection{Dataset Cleaning and Retraining}
We then removed all the epochs flagged as \textit{SUSPICIOUS} from the training set and retrained the TinySleepNet model
from scratch on the cleaned dataset using identical hyperparameters and the same experimental conditions. Out of 22,689 candidate epochs, 10,478 (46.2\%) were flagged as \textit{SUSPICIOUS} and removed, corresponding to approximately 1.4\% of total training data.

\subsection{Ablation Study}
To evaluate the proposed method's sensitivity to the confidence threshold, we repeated the pipeline with $\tau = 0.7$, which requires the LLM to run more training epochs for verification.

\section{Results}
\subsection{Baseline performance}
The reimplementation of TinySleepNet, trained on the original Sleep-EDF dataset, achieved an overall accuracy of 81.8\% and a Macro F1-Score of 75.6\% across 20-fold cross-validation. The results are slightly lower than the accuracy reported in the original TinySleepNet paper, which can be attributed to two factors: first, the original implementation of the architecture was developed in TensorFlow, whereas this study employs a PyTorch reimplementation of the architecture, which may introduce minor differences in weight initialization, optimization behavior, and numerical precision. Second, the original paper evaluated on a different subset and configuration of the sleep-EDF dataset. Despite this slight difference, the baseline results are consistent with other PyTorch-based reimplementations reported in the literature. Moreover, this study focuses on the relative improvement of the proposed LLM-assisted label-noise filtering method rather than replicating absolute baseline performance.
\subsection{LLM Filtering Statistics}
Applying the confidence threshold $\tau = 0.5$ yielded 22,689 candidate epochs out of 761,414 total training epochs across all 20 folds (approximately 3\%). Of these 22,689 candidate epochs, 10,478 (46.2\%) were flagged as \textit{SUSPICIOUS} and removed, while 12,211 (53.8\%) were detected as \textit{RELIABLE} and kept. In other words, the removed epochs represent only 1.4\% of the total training data. The proportion of candidates classified as SUSPICIOUS varied across folds, ranging from 34.7\% to 58.7\% with an average of 46.1\%.

\begin{figure}[h]
\centering
\includegraphics[width=\columnwidth]{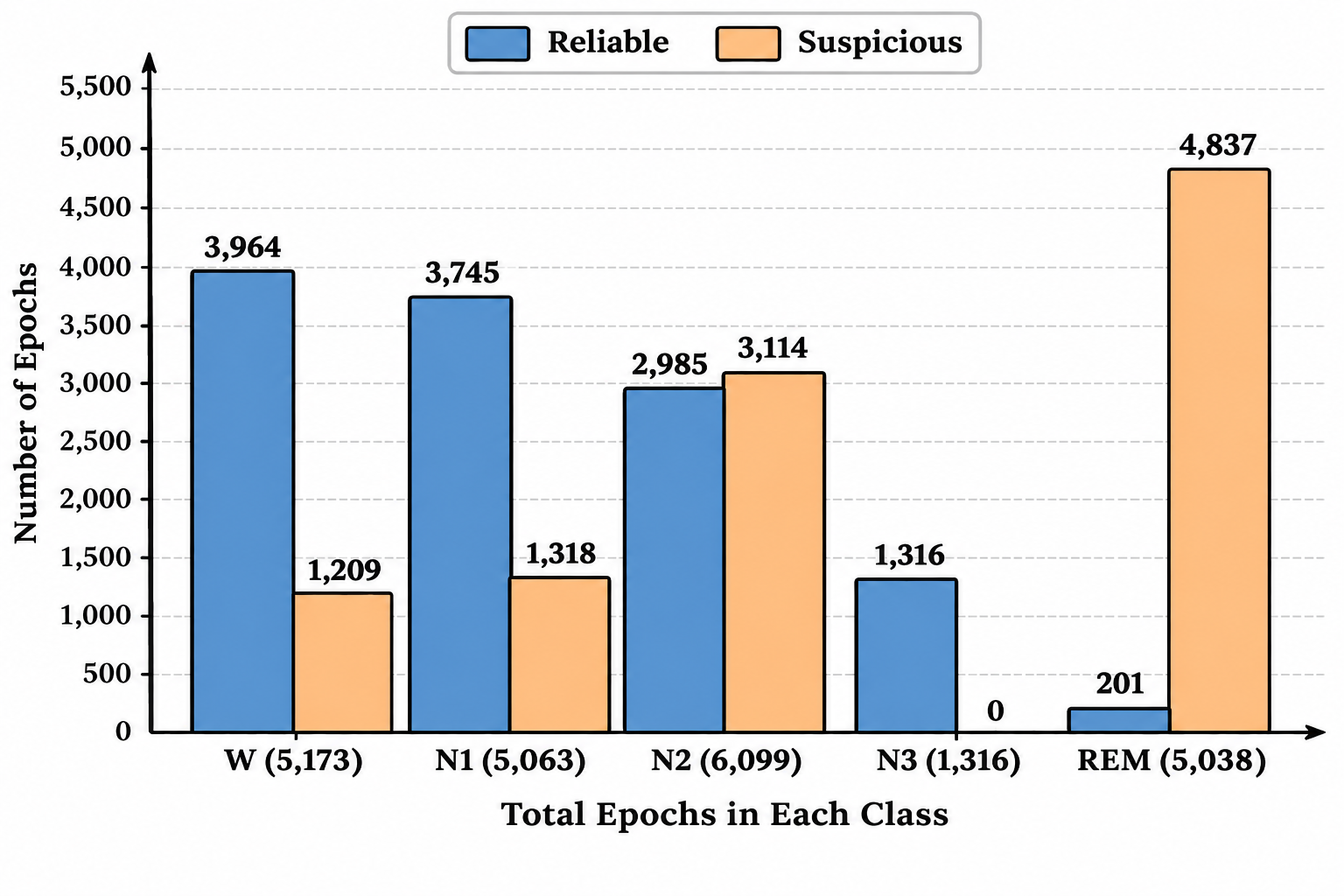}
\caption{LLM Filtering Analysis Per Stage}
\label{fig:Reliable and Suspicious Epochs by Class}
\end{figure}

Fig~\ref{fig:Reliable and Suspicious Epochs by Class} presents the distribution of LLM verdicts across sleep stages. N3 has a near-zero \textit{SUSPICIOUS} rate (0.1\%) because high delta power, its defining characteristic, is consistently present in the extracted features, making it straightforward for the LLM to verify its reliability. In contrast, REM shows the highest \textit{SUSPICIOUS} rate (96.0\%). One possible explanation is that elevated delta power in some REM epochs conflicts with the mixed-frequency EEG characteristics expected under AASM criteria, leading the LLM to classify these epochs as \textit{SUSPICIOUS}. N2 showed a \textit{SUSPICIOUS} rate of 51.1\%, potentially reflecting discrepancies between its extracted spectral features and expected sleep-stage characteristics.

\subsection{Classifier Confidence Analysis}
\begin{figure}[h]
\centering
\includegraphics[width=\columnwidth]{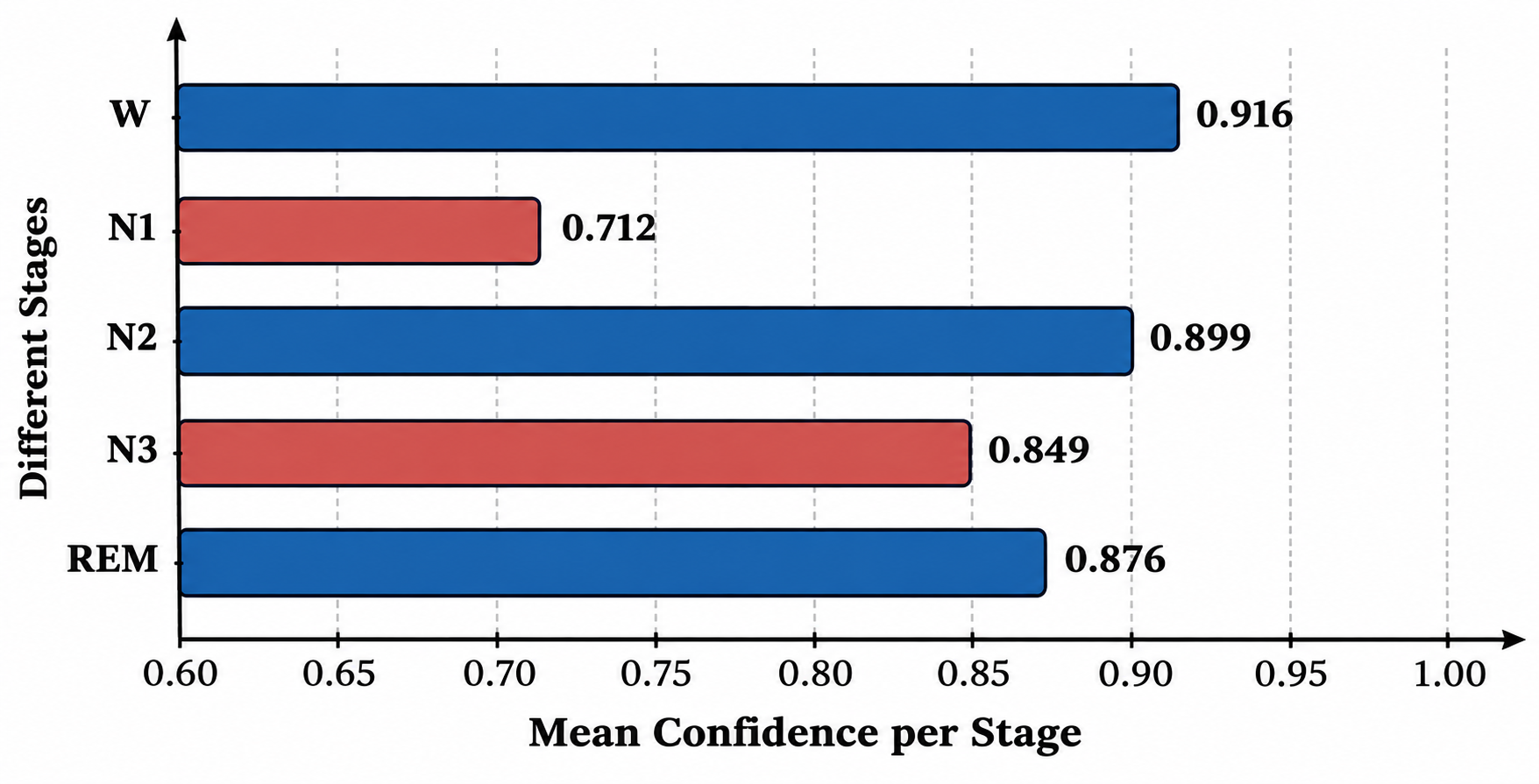}
\caption {Classifier confidence by stage (mean)}
\label{fig:Mean Confidence by Sleep Stage}
\end{figure}

According to Fig~\ref{fig:Mean Confidence by Sleep Stage}, which depicts the analysis of the per-stage mean classifier confidence scores, N1 has the lowest mean confidence of 0.712, significantly lower than all other stages (W: 0.916, N2: 0.899, REM: 0.876, N3: 0.849). This confirms that N1 is the most ambiguous stage for the classifier, consistent with its known difficulty in both automatic and manual sleep scoring. The high confidence scores for W and N2 suggest that the model has learned reliable discriminative features for these stages.

\subsection {LLM Reasoning Analysis}
Qualitative analysis of the LLM-generated reasons for \textit{SUSPICIOUS} verdicts reveals that delta power was the most frequently cited feature, mentioned in 74.5\% of suspicious reasons. This is followed by references to REM (46.4\%), sigma activity (39.3\%), and dominant frequency bands (38.3\%). For \textit{RELIABLE} verdicts, 91.5\% of reasons contained the keyword "consistent", indicating that the LLM systematically assessed the agreement between spectral features and the AASM stage criteria. These findings suggest that the LLM frequently references EEG spectral characteristics relevant to sleep-stage assessment, although independent expert evaluation is needed to confirm the clinical validity of its reasoning.

\subsection{Main Results}
After removing the \textit{SUSPICIOUS} epochs and retraining TinySleepNet from scratch on the cleaned dataset using identical hyperparameters, the proposed LLM-assisted approach achieved an overall accuracy of 84.6\% and a Macro F1-score of 77.4\%. This represents improvements 
of +2.8\% in accuracy and +1.8\% in Macro F1-score compared to the baseline. The most notable per-class improvement was in REM detection, with a +10.6\% increase in F1-score, showing that the LLM effectively identified and removed mislabeled REM epochs mistaken for N1 due to their similar mixed-frequency EEG characteristics. Wake and N2 also showed consistent improvements of +1.3\% and +1.1\%, respectively. These results confirm that the proposed LLM-based label-noise filtering method consistently improves sleep-stage classification performance without modifying the model architecture or increasing inference complexity.

\subsection{Ablation Study: Effect of Confidence Threshold}
Table~\ref{tab:ablation} compares performance across different confidence thresholds. While $\tau = 0.5$ yields the best performance with +2.8\% accuracy and +1.8\% MF1 improvement over the baseline, increasing the threshold to $\tau = 0.7$ submits significantly more epochs to the LLM (118,816 vs 22,689), resulting in more aggressive removal of 43,942 epochs (5.77\% of training data) and a performance drop below the baseline, suggesting that over-cleaning removes correctly labeled epochs.

\begin{table}[h]
\centering
\caption{Ablation Study: Effect of Confidence Threshold}
\label{tab:ablation}
\begin{tabular}{lcccc}
\hline
\textbf{Method} & \textbf{Candidates} & 
\textbf{Removed} & \textbf{Acc (\%)} & \textbf{MF1 (\%)} \\
\hline
Baseline              & -       & -      & 81.8 & 75.6 \\
$\tau$ = 0.5          & 22,689  & 10,478 & 84.6 & 77.4 \\
$\tau$ = 0.7          & 118,816 & 43,942 & 80.5 & 74.8  \\
\hline
\end{tabular}
\end{table}

\section{Discussion}
\subsection{Effectiveness of LLM-Based Noise Filtering}
Our results demonstrate that LLM-guided label-noise filtering consistently improves TinySleepNet's performance across multiple evaluation metrics. The improvement is particularly pronounced for the Wake and REM sleep stages, which have relatively clear EEG spectral signatures (alpha/beta dominance for Wake and mixed-frequency activity with absent spindles for REM). The LLM can reliably identify Wake epochs erroneously labeled as sleep stages and REM epochs confused with N2 or N1.

The more modest improvement in N2 and the slight decrease in N1 F1-score reflect the inherent ambiguity of these stages. N1 is the most challenging stage even for human experts, with inter-rater agreement as low as 50-60\%. The LLM, which reasons based on population-level AASM criteria, may remove some genuinely ambiguous N1 epochs that were correctly (if narrowly) labeled by the original annotator.

\subsection{Training-Time vs. Inference-Time LLM Usage}
A key distinguishing feature of our approach, compared to similar recent work, is that our LLM operates exclusively during training. This design choice has important practical implications: the inference pipeline retains TinySleepNet's 1.3M parameter footprint, making it suitable for deployment on edge devices, wearables, and resource-constrained clinical systems. In contrast, approaches that use large multimodal LLMs at inference time require sustained access to GPUs and incur significant per-prediction latency.

\subsection{LLM as a Second Opinion}
The LLM in our pipeline serves as a clinical second opinion rather than a replacement for annotators. Critically, when the LLM identifies a low-confidence epoch, it does not attempt to assign a corrected label. Instead, it removes the epoch from training, taking a conservative approach that avoids introducing new label errors. This is philosophically aligned with standard practice in noisy label learning, where uncertain samples are better excluded than relabeled.

\section{Conclusion}
We have presented a novel pipeline that leverages the clinical knowledge encoded in LLMs to identify and remove unreliable training labels from EEG sleep-staging datasets. By using LLaMA 3.1 as a training-time noise filter and evaluating each low-confidence epoch's EEG spectral features against AASM sleep-scoring criteria, we achieve measurable improvements in TinySleepNet's classification performance on the Sleep-EDF Expanded benchmark.

Our approach addresses a fundamental limitation of supervised sleep staging: the systematic label noise introduced by inter-rater disagreement in human annotation. Rather than accepting noisy labels as ground truth, we use the LLM's domain knowledge to conservatively remove epochs where the EEG signal characteristics are inconsistent with the assigned label. The resulting performance gains, particularly in Wake and REM classification, demonstrate that this noise removal genuinely recovers a useful training signal.

A key practical advantage of our method is that the LLM operates exclusively during training, leaving the lightweight TinySleepNet architecture unchanged at inference time. This preserves the model's suitability for real-time and edge deployment scenarios while improving the quality of its underlying training data.

Future work will explore: (1) applying this pipeline to other sleep staging architectures and datasets, (2) extending the LLM prompt to incorporate multi-channel EEG, EOG, and EMG features, (3) using LLM confidence scores for soft label weighting rather than binary filtering, and (4) validating on datasets with multiple expert annotations to directly quantify noise correction 

\section{Limitations and Future Directions}
Several limitations of the present study should be acknowledged:
Firstly, the Sleep-EDF Expanded dataset provides a single expert annotation per epoch. While we argue that our filtering improves the quality of the training data, we cannot directly verify how many removed epochs were genuinely mislabeled versus genuinely ambiguous, as no multi-annotator ground truth is available. Validation on datasets with multiple annotators per epoch (e.g., MASS) would provide stronger evidence. Secondly, recordings in the Sleep-EDF Cassette category were conducted at home, with subjects wearing EEG equipment for extended periods. This results in elevated delta power across all sleep stages due to recording artifacts and resting-state activity, thereby deviating from the strict AASM spectral criteria. The LLM prompt is calibrated to account for this through relative rather than absolute feature comparisons, but this remains a dataset-specific characteristic. Thirdly, our approach uses only the Fpz-Cz EEG channel, whereas clinical AASM staging incorporates multiple EEG channels, EOG, and EMG signals. The addition of EOG and EMG features would particularly benefit REM detection and N1 classification. Lastly, the LLM confidence score in the model's response is self-reported and may be poorly calibrated. Future work could explore using LLM confidence as a soft weighting factor rather than a binary \textit{RELIABLE/SUSPICIOUS} decision.

\vspace{12pt}


\begin{thebibliography}{00}
\bibitem{b1} M. R. Irwin, “Why sleep is important for health: A psychoneuroimmunology
perspective,” Annu. Rev. Psychol., vol. 66, pp. 143–72, 2015.

\bibitem{b2} J. Wang et al., “CareSleepNet: A Hybrid Deep Learning Network for Automatic Sleep Staging,” IEEE Journal of Biomedical and Health Informatics, pp. 1–14, Jan. 2024, doi: https://doi.org/10.1109/jbhi.2024.3426939.

\bibitem{b3} C. Iber, S. Ancoli-Israel, A. L. Chesson, and S. F. Quan, The American Academy of Sleep Medicine (AASM): Manual for the Scoring of Sleep and Associated Events: Rules, Terminology and Technical Specifications, vol. 1. Westchester, IL, USA: American Academy of Sleep Medicine, 2007.
‌
\bibitem{b4} R. S. Rosenberg and S. Van Hout, “The American Academy of Sleep Medicine inter-scorer reliability program: sleep stage scoring,” Journal of Clinical Sleep Medicine, vol. 9, no. 1, pp. 81–87, 2013, doi: 10.5664/jcsm.2350.

\bibitem{b5} H. Danker-Hopfe et al., “Interrater reliability for sleep scoring according to the Rechtschaffen \& Kales and the new AASM standard,” Journal of Sleep Research, vol. 18, no. 1, pp. 74–84, 2009, doi: 10.1111/j.1365-2869.2008.00700.x.

\bibitem{b6} E. Alickovic and A. Subasi, "Ensemble SVM method for automatic sleep stage classification," IEEE Trans. Instrum. Meas., vol. 67, pp. 1258–1265, 2018.

\bibitem{b7} A. Supratak et al., "DeepSleepNet: a model for automatic sleep stage scoring based on raw single-channel EEG," IEEE Trans. Neural Syst. Rehabil. Eng., vol. 25, pp. 1998–2008, 2017.

\bibitem{b8} H. Phan et al., "SeqSleepNet: end-to-end hierarchical recurrent neural network for sequence-to-sequence automatic sleep staging," IEEE Trans. Neural Syst. Rehabil. Eng., vol. 27, pp. 400–410, 2019.

\bibitem{b9} H. Phan, K. Mikkelsen, O. Y. Chén, P. Koch, A. Mertins, and M. De Vos, “SleepTransformer: Automatic sleep staging with interpretability and uncertainty quantification,” arXiv [cs.LG], 2021.

\bibitem {b10} K. Singhal et al., “Large language models encode clinical knowledge,” Nature, vol. 620, pp. 172–180, 2023, doi: 10.1038/s41586-023-06291-2.

\bibitem {b11} T. Yu \textit{et al.}, “SleepGPT: A Sleep Stage Language Model for Efficient Sleep Assessment,” \textit{medRxiv}, 2024. [Online]. Available: https://www.medrxiv.org/content/10.1101/2024.10.26.24316166v4

\bibitem{b12} A. B. Klok, J. Edin, M. Cesari, A. N. Olesen, P. Jennum, and H. B. D. Sorensen, "A New Fully Automated Random-Forest Algorithm for Sleep Staging," in Proc. Annual International Conference of the IEEE Engineering in Medicine and Biology Society (EMBC), 2018, pp. 4920–4923, doi: 10.1109/EMBC.2018.8513413.

\bibitem{b13} K. Yang et al., “ECG-LM: Understanding Electrocardiogram with a Large Language Model,” Health Data Science, vol. 5, Art. no. 0221, 2025, doi: 10.34133/hds.0221.

\bibitem{b14} B. Kemp et al., "Sleep-EDF Expanded Database," PhysioNet, 2000. [Online]. Available: https://physionet.org/content/sleep-edfx/1.0.0/. doi: 10.13026/C2X676

\bibitem{b15} A. Supratak and Y. Guo, “TinySleepNet: An efficient deep learning model for sleep stage scoring based on raw single-channel EEG,” in Proc. IEEE Engineering in Medicine and Biology Society (EMBC), 2020, doi: 10.1109/EMBC44109.2020.9176741.

\bibitem{b16} R. G. Norman et al., “Interobserver agreement among sleep scorers from different centers in a large dataset,” Sleep, vol. 23, no. 7, pp. 901–908, 2000.

\bibitem{b17} B. Frenay and M. Verleysen, "Classification in the Presence of Label Noise: A Survey," in IEEE Transactions on Neural Networks and Learning Systems, vol. 25, no. 5, pp. 845-869, May 2014, doi: 10.1109/TNNLS.2013.2292894.

\bibitem{b18} H. Danker-Hopfe et al., “Interrater reliability for sleep scoring according to the Rechtschaffen \& Kales and the new AASM standard,” Journal of Sleep Research, vol. 18, no. 1, pp. 74–84, 2009, doi: 10.1111/j.1365-2869.2008.00700.x.

\bibitem{b19} C. G. Northcutt, L. Jiang, and I. L. Chuang, “Confident learning: Estimating uncertainty in dataset labels,” Journal of Artificial Intelligence Research, vol. 70, pp. 1373–1411, Apr. 2021.

\bibitem{b20} A. Blum and T. Mitchell, “Combining labeled and unlabeled data with co-training,” in Proc. 11th Annu. Conf. Computational Learning Theory (COLT), Madison, WI, USA, 1998, pp. 92–100.

\bibitem{b21} R. A. Jacobs, M. I. Jordan, S. J. Nowlan, and G. E. Hinton, “Adaptive mixtures of local experts,” Neural Computation, vol. 3, no. 1, pp. 79–87, Mar. 1991.

\bibitem{b22}  K. Singhal et al., “Towards expert-level medical question answering with large language models,” arXiv:2305.09617, 2023.

\bibitem{b23} K. Singhal et al., “Towards expert-level medical question answering with large language models,” arXiv:2305.09617, 2023.

\bibitem{b24} S. Zhao, M. Peng, H. Jiang, T. Li, and S. Li, “EEG Agent: A unified framework for automated EEG analysis using large language models,” in Proc. AAAI Conf. Artificial Intelligence, vol. 40, no. 21, pp. 18063–18071, Mar. 2026.

\bibitem{b25} X. Qiu, C. Zhan, G. Ma, J. Huang, and X. Tan, “Robust sleep stage prediction from electroencephalogram with label noise using multimodal large language models,” in Proc. Int. Conf. Medical Image Computing and Computer-Assisted Intervention (MICCAI), Cham, Switzerland: Springer, pp. 575–585, Sep. 2025.

\bibitem{b26} A. Rechtschaffen and A. E. Kales, A manual of standardized terminology, techniques and scoring systems for sleep stages of human subjects. Los Angeles, CA, USA: UCLA Brain Information Service, Brain Research Institute, 1968.

\end{thebibliography}
\end{document}